# USER INTENT RECOGNITION AND SEMANTIC CACHE OPTIMIZATION-BASED QUERY PROCESSING FRAMEWORK USING CFLIS AND MGR-LAU

Sakshi Mahendru

*Abstract*—Query Processing (QP) is optimized by a Cloud-based cache by storing the frequently accessed data closer to users. Nevertheless, the lack of focus on user intention type in queries affected the efficiency of QP in prevailing works. Thus, by using a Contextual Fuzzy Linguistic Inference System (CFLIS), this work analyzed the informational, navigational, and transactional-based intents in queries for enhanced QP. Primarily, the user query is parsed using tokenization, normalization, stop word removal, stemming, and POS tagging and then expanded using the WordNet technique. After expanding the queries, to enhance query understanding and to facilitate more accurate analysis and retrieval in query processing, the named entity is recognized using Bidirectional Encoder UnispecNorm Representations from Transformers (BEUNRT). Next, for efficient QP and retrieval of query information from the semantic cache database, the data is structured using Epanechnikov Kernel-Ordering Points To Identify the Clustering Structure (EK-OPTICS). The features are extracted from the structured data. Now, sentence type is identified and intent keywords are extracted from the parsed query. Next, the extracted features, detected intents and structured data are inputted to the Multi-head Gated Recurrent Learnable Attention Unit (MGR-LAU), which processes the query based on a semantic cache database (stores previously interpreted queries to expedite effective future searches). Moreover, the query is processed with a minimum latency of 12856ms. Lastly, the Semantic Similarity (SS) is analyzed between the retrieved query and the inputted user query, which continues until the similarity reaches 0.9 and above. Thus, the proposed work surpassed the previous methodologies.

I. INTRODUCTION

The execution of database commands that includes parsing, optimization, and execution, to retrieve information efficiently is termed the QP [1]. To expedite effective future searches, the usage of semantic cache in QP stores previously interpreted queries [2,3]. However, QP suffers from slower performance [4], inefficient resource usage, and difficulty in handling complex queries effectively [5]. Thus, for improved analysis, the enhancement of QP is needed.

For query optimization, prevailing techniques like the deep learning-based GRU method [6] were primarily employed [7]. For enhancing QP, the machine learning-based Support Vector Machine classified queries [8,9,10]. Nevertheless, the user-intentions of queries that degraded the accurate QP were not concentrated by these techniques. Thus, to improve QP efficiency and accuracy, the proposed work integrates various techniques.

A    Problem Statement

The limitations of prevailing works are described below,

➢ In prevailing works, neglected user intentions in queries hampered accurate QP.

➢ Due to language nuances, contextual variations, Named Entity Disambiguation, structural differences, and high computational resource demands, measuring SS was complex in prevailing works. These factors hindered accurate query-data matching.

➢ [11] lacked the recognition and disambiguation of named entities that impacted the accurate information retrieval.

➢ Difficulty in converting the queries into structured representations in (Qiu et al., [12]), limited the efficiency of query execution.

The objectives of the proposed work are detailed below,

- ➢ By utilizing the CFLIS technique, the user-intent queries are detected for accurate QP.

- ➢ By using CS, SS between the retrieved query and inputted user query is analyzed for enhanced QP.

- ➢ Disambiguation of named entities enhances accurate information retrieval.

- ➢ The queries are converted to structured data using EK-OPTICS for efficient query execution.

The remaining part is arranged as: the associated works are elucidated in Section 2; the proposed methodology is described in Section 3; the results and discussion are presented in Section 4; the paper is concluded with the future development in Section 5.

## II. LITERATURE SURVEY

Ahmad et al., [11] presented a Semantic Cache QP framework to enhance the QP efficiency. For improved QP performance, this work used query dispensation algorithms and cache management techniques along with the incorporation of data mining and data warehousing. But, cache utilization was affected by the inaccurate identification of overlapping data.

Qiu et al., [12] advanced a decentralized QP technique for aggregated Boolean queries. Here, Boolean queries were decomposed with global sets that were broadcasted across edge servers. For query acceleration, this technique employed a tree-based method. However, due to synchronization overhead, managing global sets across edge servers created scalability issues, hindering the QP efficiency.

Akhtar et al., [13] developed an Adaptive Cache Replacement (ACR) system that enhanced QP in linked open data cloud by caching frequently accessed query results. Here, based on edit distance, ACR identified similar queries. Despite enhanced QP, QP accuracy was degraded by the non-identification of semantic relationships between query and cached data.

Michiardi et al., [14] merged in-memory caching and multi-query optimization to enhance QP's efficiency. For solving the multiple-choice knapsack problem, the execution plan was optimized for QP enhancement. But, the QP was enhanced without acknowledging the query intent, affecting the overall performance of the model.

Chang et al., [15] developed the oblivious query framework using Oblivious Random Access Memory (ORAMs) for efficient QP over cloud databases. To process K-Nearest Neighbor queries with high throughput, this work employed ORAM caching and batch processing. Yet, this framework enhanced QP without recognizing and disambiguating the named entities, which reduced the accurate information retrieval process in QP.

## III. PROPOSED METHODOLOGY FOR ENHANCED QUERY PROCESSING

In Figure 1, the CFLIS AND MGR-LAU-based QP is depicted.

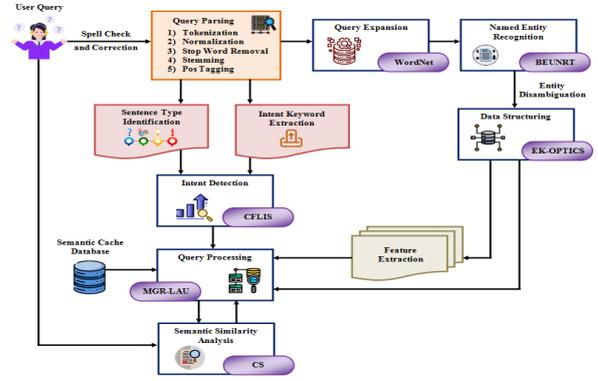

Fig. 1: Structural Diagram of the Proposed Work

### A    Query Collection and Parsing

The proposed work starts by collecting and preprocessing the user queries. Here, the $(n)$ numbers of collected queries $(C^{queries})$ are parsed as mentioned below,

- Initially, $(C^{queries})$ are tokenized $(C^{tok})$ using the tokenization process $(T_\kappa)$ as shown in below equation,

$$C^{tok} = T_\kappa(C^{queries}) \quad (1)$$

- Next, for effective analysis, $(C^{tok})$ queries are normalized using lowercasing, punctuation removal, handling contractions, and abbreviations. Hence, the $(s)$ numbers of normalized queries $(C^{norm})$ are given as,

$$C^{norm} = C^1, C^2, \ldots\ldots, C^s \quad (2)$$

- Then, the stop words are removed and the process of removing stop words $(\vartheta)$ is formulated as,

$$C^{SR} = \vartheta(C^{norm}) \quad (3)$$

Where, $(C^{SR})$ signifies the filtered query after the removal of stopwords. Next, $(C^{SR})$ are stemmed and represented as $(C^{stem})$.

- Finally, the POS is tagged in $(C^{stem})$ to represent the grammatical structures as shown below,

$$C^{pos} = \vartheta^{tagger}(C^{stem}) \quad (4)$$

Where, $(\vartheta^{tagger})$ signifies the tagger function that assigns the POS to $(C^{stem})$ and $(C^{pos})$ represents the POS tagged queries.

Hence, the query is parsed based on the above-mentioned tokenization, normalization, stop word removal, stemming, and POS tagging, and the finally parsed query output is represented as $(C^{parsed})$.

### B    Query Expansion

Next, $(C^{parsed})$ is inputted to WordNet, which is a lexical database that organizes words into synsets or sets of synonymous words and describes relationships betwixt them. The specific relationship is calculated by measuring the semantic similarity between two synsets (words) in $(C^{parsed})$ by comparing their Information Content (IC). Hence, $(C^{relation})$ computes the ratio of the IC's Least Common Subsumer $(v^n)$ to the average IC $(e^n)$ of the individual synsets. Based on this, the specific relationship $(C^{relation})$ between two words $(v^n, e^n)$ in $(C^{parsed})$ is given as,

$$C^{relation} = 2 * \frac{C^{parsed}(I^{content})}{C^{parsed}\left(\sum_{n=1}^{j}(v+e)\right)} \quad (5)$$

Where, the information content in $(C^{parsed})$ is signified as $(I^{content})$, and the total number of similar words in $(C^{parsed})$ is depicted as $(j)$. Thus, the query is expanded based on $(C^{relation})$ and is denoted as $(Q^{ex})$.

### C    Named Entity Recognition and Entity Disambiguation

By extracting contextual features, $(Q^{ex})$ is then inputted to BERT for Named Entity Recognition (NER). But, the NER processing is slowed by the BERT's extensive weight update. Thus, UnispecNorm (UN) is used that uniformly initialize the weights by applying spectral normalization for reduced processing time. The algorithmic steps of BEUNRT are detailed below as,

→ Initially, by using the embedded matrix $(M_{EM})$, $(Q^{ex})$ is converted into the embedded vector $(V_{EM})$, which is given as,

$$V_{EM} = M_{EM}(Q_{ex}) \quad (6)$$

→ Here, to reduce the duration of processing time, the weights in each step are initialized using the UN method, which is equated as,

$$w^* = \frac{U \sim w(a,b)}{\varsigma(w)} \quad (7)$$

Here, the initialized weights is signified as $(w^*)$, the uniform distribution function is represented as $(U)$, the lower and upper bounds of $(U)$ is depicted as $(a,b)$, $(w)$ represents the weight matrix, and $(\varsigma)$ represents the largest singular value of $(w)$.

→ Then, the Self Attention Mechanism $(A^{self})$ is applied using query $(Q)$, key $(K)$, and vector $(V)$ with dimension $(d_K)$ as,

$$A^{self} = \zeta\left(\frac{QK^{T'}V}{\sqrt{d_K}}\right) \quad (8)$$

Where, $(T')$ signifies the transpose function and $(\zeta)$ denotes the softmax activation that is expressed as,

$$\zeta(A^{self}) = \frac{e^{A^{self}}}{\Sigma e^{A^{self}}} \quad (9)$$

Where, $(e)$ signifies the exponential function. To recognize the named entities, the output $(A^{self})$ is then pooled $(A^P)$ and fine-tuned $(F^{tunned})$, which are denoted as $(R^{NE})$.

Next, to resolve ambiguity in named entities, the entity disambiguation is done. Hence, the correctly identified named entities are represented as $(C^{NER})$.

### D  Data Structuring and Feature Extraction

Next, to make the QP easier, $(C^{NER})$ is structured along with the Hadoop Distributed File System (HDFS) employing Ordering Points To Identify the Clustering Structure (OPTICS). Nevertheless, suboptimal clustering structures are caused by the random selection of Minimum Points $(MinPts)$ and an additional parameter called epsilon $(\varepsilon'')$. Thus, the Epanechnikov Kernel (EK) is utilized instead of random selection to obtain the optimal solutions, which is detailed below

- ♣ Initially, the $(z)$ numbers of $(C^{NER})$ are initialized as,

$$C^{NER} = C^1, C^2, \ldots, C^z \quad (10)$$

- ♣ Next, to obtain the more reliable $(MinPts)$ and $(\varepsilon'')$ values, the density estimation is done for each $(C^{NER})$ using the EK function and is given as,

$$E^k = \frac{3}{4}\left(1 - (C^{NER})^2\right) \text{ for } |C^{NER}| \leq 1 \quad (11)$$

Where, $|C^{NER}|$ signifies the absolute data of $(C^{NER})$ and $(E^k)$ denotes the estimated density.

- ♣ From $(E^k)$, the $(MinPts)$ and $(\varepsilon'')$ are selected using the below formula as,

$$MinPts = \alpha \cdot \max(E^k) \quad (12)$$

$$\varepsilon'' = \min\{\widetilde{d} \mid E^k < \beta \cdot \max(E^k)\} \quad (13)$$

Where, $(\alpha)$ signifies the scaling factor, $(\beta)$ denotes the constant, and $(\widetilde{d})$ embodies the Euclidean Distance. Now, the data is structured by using $(MinPts \text{ and } \varepsilon'')$ and denoted as $(S^D)$.

From $(S^D)$, the features, such as Term Frequency-Inverse Document Frequency (TF-IDF), numerical values, categorical attributes, temporal data, and relationships between entities are extracted and denoted as $(E^{features})$.

### E  Intent Keyword Extraction and Sentence Type Identification

For now, the information intent keywords (such as "how to", "what is", and "definition of"), navigational intent keywords (such as brand names, product names, specific URLs, and terms like "login" and "download"), and transactional intent keywords (such as "buy", "purchase", "checkout", "subscribe", and "membership") are extracted from $(C^{parsed})$ and denoted as $(K^{features})$. The explanation of keywords is provided below as,

a. The users' desire to acquire knowledge or find answers to specific questions is termed the information intent. They seek to understand topics or solve problems, often using queries like "how to" or "what is," aiming for informative, detailed, and accurate content.

b. A specific type of search query, where users seek a particular website, brand, or web page is termed the navigational intent. It is applicable when users already have a destination in mind and use search engines as a tool to reach that destination quickly. For example, keywords like "Starbucks," "Gmail login," or "download" exhibit navigational intent because users are looking for specific online locations or platforms.

c. The search intent, where users actively seek to make a purchase or take immediate action is termed the transactional intent. Also, the transactional search intent is characterized by users who are ready to make a purchase or engage in a specific transaction. Therefore, keywords related to buying, such as "buy," "purchase," "order," or "get," are commonly associated with transactional intent. Therefore, the transactional intent is crucial for e-commerce and services to optimize content for such intent.

Moreover, the sentence types, such as statement, interrogative, imperative, and more are identified from $\left(C^{parsed}\right)$ which are denoted as $\left(S^{type}\right)$. These are done to enhance intent detection.

### F  Intent Detection

The intent is detected from $\left(K^{features}, S^{type}\right)$ using the Fuzzy Inference System (FIS) to get the retrieved information accurately. But, due to simplistic membership functions, conventional FIS may lack accuracy. Thus, for more nuanced and contextually appropriate fuzzy reasoning, the Contextual Linguistic Membership Function (CLMF) is utilized. The working steps of CFLIS are described further,

Rule Generation

Initially, the rules $\left(R^F\right)$ are set based on the if-then condition as shown below,

$$R^F = \begin{cases} if \ K^{feaures} \in A^1 \ then \ I \\ if \ K^{feaures} \in A^2 \ then \ N \\ if \ K^{feaures} \in A^3 \ then \ T \end{cases} \quad (14)$$

Where, the interrogative, statement, and imperative sentences are depicted as $\left(A^1, A^2, A^3\right)$. The condition states that if $\left(K^{features}\right)$ belongs-to $\left(A^1\right)$, then informational intent $(I)$ will be detected, if $\left(K^{features}\right)$ belongs-to $\left(A^2\right)$, then navigational intent $(N)$ will be detected, and if $\left(K^{features}\right)$ belongs-to $\left(A^3\right)$, then transactional intent $(T)$ will be detected.

Membership Function

Next, the CLMF membership function $\left(L^M\right)$ is assigned for a fuzzy set that is given as,

$$L^M = f\left(R^F, c^\varphi\right) \times \left(\frac{1}{\sigma\sqrt{2\pi}} e^{-\left(\frac{\left(R^F-\mu\right)^2}{2\sigma^2}\right)}\right) \quad (15)$$

Where, $f\left(R^F, c^\varphi\right)$ signifies the function $(f)$ that captures contextual influence $\left(c^\varphi\right)$ based on $\left(R^F\right)$, and $(\mu, \sigma)$ denotes the mean and standard deviation of $\left(R^F\right)$, and $(\pi = 22/7)$.

Fuzzification

Next, based on the fuzzy relationship $\left(F^{rel}\right)$, the rules are fuzzified as shown below,

$$D^B = \frac{\sum\left(K^{features}, S^{type}\right) \times F^{rel}\left(R^F\right)}{L^M} \quad (16)$$

Where, $\left(D^B\right)$ signifies the final decision obtained using $\left(F^{rel}\right)$.

Defuzzification

To obtain the crisp output $\left(p^{crisp}\right)$, the fuzzified output is then defuzzified.

$$p^{crisp} = \frac{\sum D^B \times L^M}{\sum L^M} \quad (17)$$

**Pseudo code of CFLIS**

**Input:** Extracted Keywords, and Type sentence, $\left(K^{features}, S^{type}\right)$
**Output:** Detected intent, $\left(Q^{in}\right)$
**Begin**
    **Initialize** iterations $\left(t, t^{max}\right)$
    **While** $\left(t < t^{max}\right)$
        **Set** rules, $\left(R^F\right)$
        **Evaluate** $\left(L^M\right)$,

$$L^M = f\left(R^F, c^\varphi\right) \times \left(\frac{1}{\sigma\sqrt{2\pi}} e^{-\left(\frac{\left(R^F-\mu\right)^2}{2\sigma^2}\right)}\right)$$

        **Fuzzify** the rules,

$$D^B = \frac{\sum\left(K^{features}, S^{type}\right) \times F^{rel}\left(R^F\right)}{L^M}$$

        **Obtain** $\left(p^{crisp}\right)$
    **End** while
    **Return** $\rightarrow \left(Q^{in}\right)$
**End**

Therefore, the user's intent is classified as informational, navigational, and transactional based on rules and is denoted as $(Q^{in})$.

### G  Query Processing

Lastly, to retrieve information, the extracted features, structured data, and detected intents are inputted to the GRU classifier as $(\delta^\varepsilon)$. Nevertheless, GRU has a limited capacity for complex pattern recognition. Therefore, to enhance pattern recognition, Multi-Head Attention with Learnable Attention Heads (MHA-LAH) is utilized in GRU for improved QP efficiency. In Figure 2, the MGR-LAU classifier is depicted.

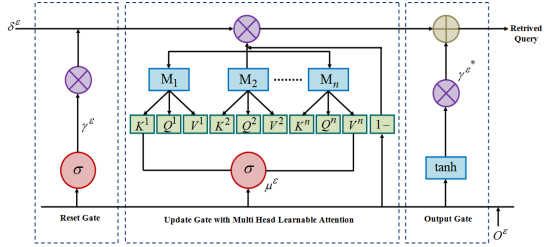

Fig. 2: MGR-LAU classifier

→ Primarily, the reset $(r^\varepsilon)$ and update $(u^\varepsilon)$ gates with time $(\varepsilon)$ are calculated from input $(\delta^\varepsilon)$,

$$r^\varepsilon = \sigma\left(w^{r^\varepsilon}.\delta^\varepsilon + w^{r^\varepsilon}.\gamma_{\varepsilon-1}\right) \quad (18)$$

$$u^\varepsilon = \sigma\left(w^{u^\varepsilon}.\delta^\varepsilon + w^{u^\varepsilon}.\gamma_{\varepsilon-1}\right) \quad (19)$$

Where, the weights for $(r^\varepsilon, u^\varepsilon)$ gates is depicted as $(w^{r^\varepsilon}, w^{u^\varepsilon})$, the hidden state with time $(\varepsilon)$ is signified as $(\gamma)$, and the sigmoidal activation is depicted as $(\sigma)$, which is given as,

$$\sigma(\delta^\varepsilon) = \frac{1}{1+e^{-\delta^\varepsilon}} \quad (20)$$

→ Next, to diverse query aspects in semantic cache systems, MHA-LAH enhances pattern recognition. So, the concatenation formula using MHA and LAH is equated from $(A^{self})$ as,

$$A_v^{self}(QW_{Qv}, KW_{Kv}, VW_{Vv}) = \lambda(\gamma_\varepsilon^1, \gamma_\varepsilon^2 \ldots \ldots \gamma_\varepsilon^r) L_\varpi \quad (21)$$

Where, $(W_{Qv}, W_{Kv}, W_{Vv})$ signifies the weights of $(Q, K, V)$ in $(v^{th})$ attention head, $(\lambda)$ denotes the concatenation operation, $(\gamma_\varepsilon^1, \gamma_\varepsilon^2 \ldots \ldots \gamma_\varepsilon^r)$ signifies the $(r)$ numbers of $(\gamma_\varepsilon)$, and $(L_\varpi)$ represents the transformation matrix.

→ Then, $(\gamma)$ is updated $(\gamma^*)$ with $(\tanh)$ activation, which is computed as,

$$\gamma_\varepsilon^* = \tanh\left(w^{\gamma^\varepsilon}.\delta^\varepsilon + r^\varepsilon \circ \left(w^{\gamma^\varepsilon}.\gamma_{\varepsilon-1}\right)\right) \quad (22)$$

$$\tanh(\delta^\varepsilon) = \frac{\left(e^{\delta^\varepsilon} - e^{-\delta^\varepsilon}\right)}{\left(e^{\delta^\varepsilon} + e^{-\delta^\varepsilon}\right)} \quad (23)$$

→ The updated hidden state $(\gamma^*)$ is iterated in the output gate $(o^\varepsilon)$, which is given as,

$$o^\varepsilon = (1-u^\varepsilon) \circ \gamma_{\varepsilon-1} + r^\varepsilon \circ \gamma^* \quad (24)$$

Hence, the query is retrieved from $(o^\varepsilon)$ and is represented as $(R^{query})$.

**Pseudo code of MGR-LAU**

**Input:** Combined input, $(\delta^\varepsilon)$
**Output:** Retrieved Query, $(R^{query})$
**Begin**
  **Initialize** iterations, $(\varepsilon, \varepsilon^{\max})$
  **While** $(\varepsilon, < \varepsilon^{\max})$
    **Initialize** $(\delta^\varepsilon)$
    **Compute** $(r^\varepsilon), (u^\varepsilon)$
    **Concatenate** MHA-LAH,

$$A_v^{self}(QW_{Qv}, KW_{Kv}, VW_{Vv}) = \lambda(\gamma_\varepsilon^1, \gamma_\varepsilon^2 \ldots \ldots \gamma_\varepsilon^r) L_\varpi$$

  **Update** $(\gamma^*)$,
$$\gamma_\varepsilon^* = \tanh\left(w^{\gamma^\varepsilon}.\delta^\varepsilon + r^\varepsilon \circ \left(w^{\gamma^\varepsilon}.\gamma_{\varepsilon-1}\right)\right)$$
  **Evaluate**
$$o^\varepsilon = (1-u^\varepsilon) \circ \gamma_{\varepsilon-1} + r^\varepsilon \circ \gamma^*$$
  **End** while

$$\text{Return} \rightarrow \left(R^{query}\right)$$
**End**

Then, the similarity is checked between $\left(R^{query}\right)$ and $\left(C^{queries}\right)$ for enhanced analysis.

### H  Semantic Similarity Analysis

Here, SS analysis is performed which ensures that the retrieved results closely match the user query using CS. The equation for SS analysis $\left(S^{\cos}\right)$ between $\left(R^{query}\right)$ and $\left(C^{queries}\right)$ is given as,

$$S^{\cos} = \frac{R^{query} \cdot C^{queries}}{\left\|R^{query}\right\| \cdot \left\|C^{queries}\right\|} \qquad (25)$$

Hence, it will be considered as an effective outcome if the similarity is higher than 0.9; otherwise, the QP continues until the similarity reaches 0.9 or above. The performance assessment of the proposed work is given below.

## IV. RESULTS AND DISCUSSION

Here, the proposed framework's performance is analogized to the prevailing methodologies. The results were obtained by working on the PYTHON platform.

### A  Dataset Description

The Centre for Inventions and Scientific Information (CISI) dataset, which is gathered from the Kaggle Platform, was used by the propose work. 1,460 documents with unique Identifications (IDs), titles, authors, abstracts, and cross-references, 112 queries with unique IDs and text, and a mapping file linking query IDs to document IDs are the 3 files encompassed in the CISI dataset. Hence, the file comprises the "ground truth" that links queries to documents for efficient information retrieval. 80% and 20% of the query are used for training and testing.

### B  Performance Analysis

Here, the performance of the proposed methods like MGR-LAU, CFLIS, EK-OPTICS, and BEUNRT is explained and compared with traditional models.

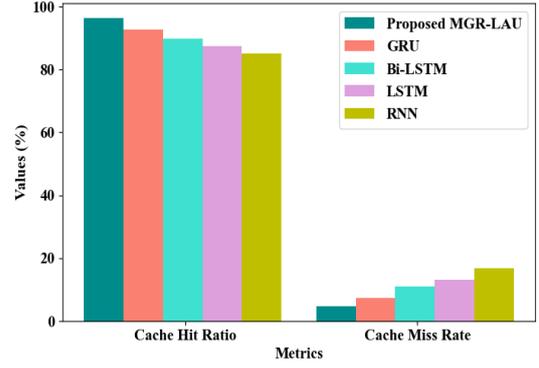

Fig. 3: Comparative Analysis for MGR-LAU

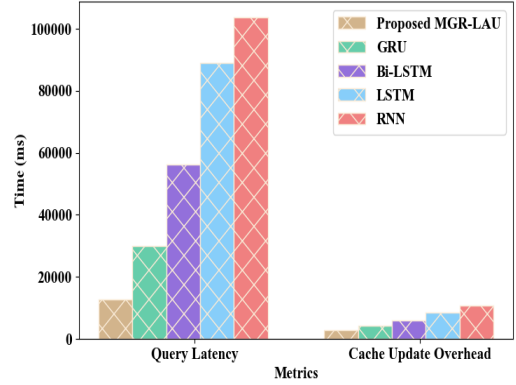

Fig. 4: Graphical Comparison of Proposed Classifier

The performance assessment of the proposed MGR-LAU technique and the existing GRU, Bidirectional Long Short Term Memory (Bi-LSTM), Long Short Term Memory (LSTM), and Recurrent Neural Network (RNN) are represented in figures 3 and 4. For capturing intricate patterns in query sequences, the proposed model used MHA-LAH. the QP was done with 96.25%, 4.85%, 12856ms, and 2896ms for Cache Hit Ratio (CHR), Cache Miss Rate (CMR), Query Latency (QL), and Cache Update Overhead (CUO), respectively. However, the existing techniques attained a minimum average CHR of 88.74% and a maximum average CMR of 12.07%, 69667ms QL, and 7289ms CUO than the proposed model. Thus, the proposed classifier's QP is better when weighed against the prevailing models.

TABLE 1: COMPARISON OF MEMORY CONSUMPTION

| Techniques | Memory Consumption (MB) |
|---|---|
| Proposed MGR-LAU | 598 |
| GRU | 924 |
| Bi-LSTM | 1279 |
| LSTM | 1867 |
| RNN | 2597 |

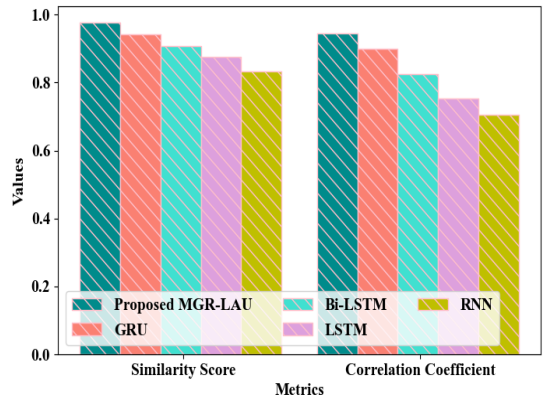

Fig. 5: Comparative Analysis of MGR-LAU

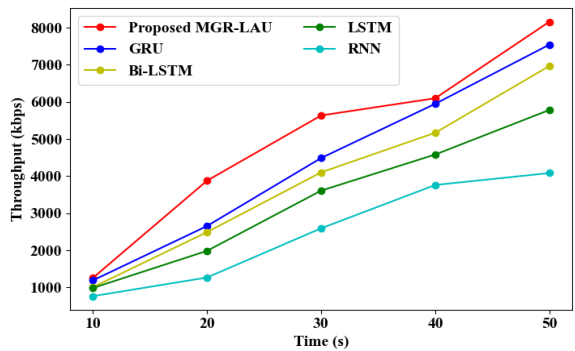

Fig. 6: Graphical Comparison Related to Throughput vs. Time

The proposed classifier captured the intricate relationships more efficiently than the existing techniques as MHA-LHA focuses on capturing syntactic structure, semantic content, and temporal dependencies in search queries. When analogized to the prevailing methodologies, the proposed model processed the query with a lower memory consumption of 598 MegaByte (MB), higher Similarity Score (SS) of 0.976 and Correlation Coefficient (CC) of 0.9454, and higher throughput of 3875kbps and 8159kbps for the 20s and 50s, respectively, which is depicted in Table 1 and Figures 5 and 6. Hence, the proposed model outperformed traditional techniques in QP.

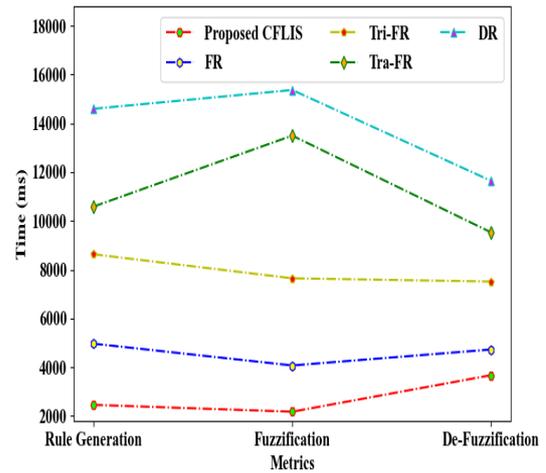

Fig. 7: Comparative Analysis of CFLIS

In Figure 7, regarding the metrics like Rule Generation Time (RGT), Fuzzification Time (FT), and De-Fuzzifictaion Time (DFT), the proposed CFLIS and the existing FR, Triangular Fuzzy Rule (Tri-FR), Trapezoidal Fuzzy Rule (Tra-FR), and Decision Rule (DR) are compared. The CLMF used as a membership function in the proposed model could identify the nuances of linguistic terms in the query. Thus, the proposed model detected the type of intent with 2459ms RGT, 2179ms FT, and 3674ms DFT, whereas the existing techniques attained higher RGT, FT, and DFT than the proposed model. Thus, when analogized to the prevailing methodologies, the proposed CFLIS identified the type of pattern in the search query more efficiently.

TABLE 2: COMPARATIVE ANALYSIS OF EK-OPTICS

| Methods | Silhouette Score |
|---|---|
| Proposed EK-OPTICS | 0.987 |
| OPTICS | 0.953 |
| DBSCAN | 0.928 |
| KMA | 0.919 |
| FCM | 0.895 |

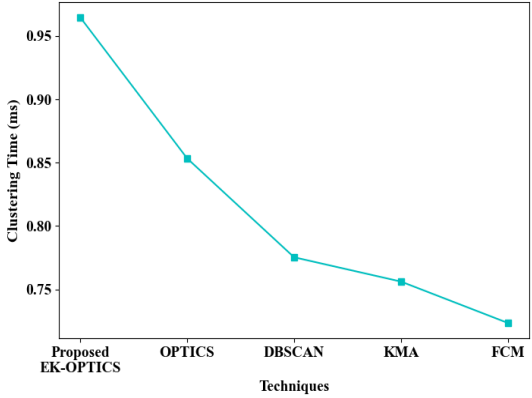

Fig. 8: Comparative Analysis Regarding Clustering Time

The proposed EK-OPTICS first selected the MinPts and epsilon using EK to structure the search query and then clustered the data along with Hadoop. Hence, the proposed model obtained a Silhouette Score of 0.987 and Clustering Time (CT) of 5219ms; but, the existing OPTICS, Density-Based Spatial Clustering of Applications with Noise (DBSCAN), K-Means Algorithm (KMA), and Fuzzy C-Means (FCM) clustering model achieved a Silhouette Score of 0.953, 0.928, 0.919, and 0.895, and CT of 9074ms, 12847ms, 17397ms, and 21237ms, respectively. Thus, when weighed against the prevailing techniques, the proposed model structured the data better.

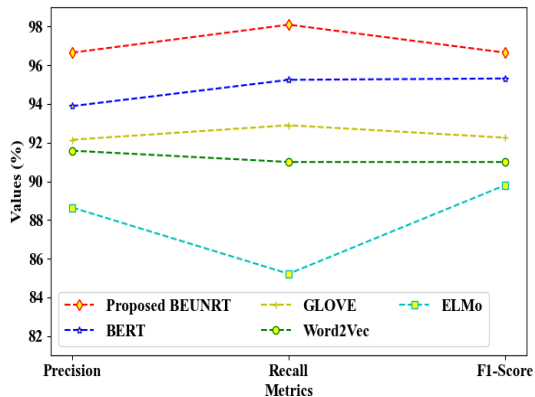

Fig. 9: Graphical Comparison for BEUNRT

The proposed BEUNRT is used for NER from the expanded query. Here, for weight initialization, the UN technique is used. Thus, regarding precision, recall, and F1-Score, the NER was done by the proposed model with 96.654%, 98.098%, and 96.654%; but, the prevailing techniques acquired (93.889%, 95.246%, 95.317%) for BERT, (92.149%, 92.898%, 92.254%) for Global Vector for Word Representation (GLOVE), (91.58%, 90.997%, 90.997%) for Word2Vec, and (88.639%, 85.214%, 89.793%). Thus, the proposed model outperformed prevailing techniques in NER.

TABLE 3: COMPARISON OF RELATED WORKS

| Study | Method | QL (ms) | Drawback |
|---|---|---|---|
| Proposed Work | MGR-LAU | 12856 | - |
| Salehpour & Davis, [16] | SymphonyDB | 52750 | The varying characteristics of the query could not be identified. |
| Cuzzocrea et al., [17] | SFS | - | The process was complex and cost-effective. |
| Shang et al., [18] | GvdsSQL | 28050 | The similarity of the query sentence was difficult to identify. |
| Sharma et al., [19] | CDSS | 35500 | Named entities could not be detected. |
| Bok et al., [20] | MSP | 274500 | Handling of QP was difficult. |

Here, based on the semantic cache, the proposed work processed the search query. Primarily, the user's search query is preprocessed. Next, NER is done using BEUNRT with a recall of 98.098% for the expanded query. Then, EK-OPTICS structured the query with a CT of 5219ms from which features were extracted. Now, the sentence type is identified from the preprocessed data. The intent is identified with an RGT of 2459ms using CFLIS. Then, the detected intent, structured data, and extracted features were given to the MGR-LAU classifier, which processed the query with a CHR of 96.25% and QL of 12856ms. Lastly, the similarity analysis was made using CS. Also, the query process continued until a similarity of 0.9 or above was attained. Hence, the proposed model processed the user's search query efficiently

V. CONCLUSION

Here, based on semantic cache, the proposed work processed the search query. Primarily, the user's search query is preprocessed. Next, NER is done using BEUNRT with a recall of 98.098% for the expanded query. Then, EK-OPTICS structured the query with a CT of 5219ms from which features were extracted. Now, sentence type is identified from the preprocessed data. The intent is identified with an RGT of 2459ms using CFLIS. Then, the detected intent, structured data, and extracted features were given to the MGR-LAU classifier, which processed the query with a CHR of 96.25% and QL of 12856ms. Lastly, the similarity analysis was made using CS. Also, the query process continued until a similarity of 0.9 or above was attained. Hence, the proposed model processed the user's search query efficiently

*Future Scope*

The query's sentence complexity, the semantic relationship between words in a sentence, and the exact meaning of the sentence were not considered despite effectively performing semantic cache-based QP. Thus, to improve the proposed work's performance, the above-mentioned process will be concentrated in the future.